\documentclass[10pt,twocolumn,letterpaper]{article}

\usepackage{cvpr}              
\usepackage[accsupp]{axessibility}

\definecolor{cvprblue}{rgb}{0.21,0.49,0.74}
\usepackage[pagebackref,breaklinks,colorlinks,allcolors=cvprblue]{hyperref}
\usepackage{listings}
\usepackage{marvosym}

\def\paperID{2120} 
\def\confName{CVPR}
\def\confYear{2025}

\newcommand{\method}{InteractAnything\xspace}

\begin{document}

\title{InteractAnything: Zero-shot Human Object Interaction Synthesis via LLM Feedback and Object Affordance Parsing}

\author{%
Jinlu Zhang$^{1,2}$, Yixin Chen$^{2\,\textrm{\Letter}}$, Zan Wang$^{2,3}$, Jie Yang$^{2,4}$,
Yizhou Wang$^{1,5,6,7\,\textrm{\Letter}}$, Siyuan Huang$^{2\,\textrm{\Letter}}$\\
\small $^\textrm{\Letter}$ indicates corresponding authors\quad{}\\
\small $^1$ Center on Frontiers of Computing Studies, School of Computer Science, Peking University\quad{}\\
\small $^2$ State Key Laboratory of General Artificial Intelligence, BIGAI\quad{}
\small $^3$ Beijing Institute of Technology\quad{}\\
\small $^4$ The Chinese University of Hong Kong, Shenzhen\quad{}
\small $^5$ Nat'l Eng. Research Center of Visual Technology, Peking University\quad{}\\
\small $^6$ Institute for AI, Peking University\quad{}
\small $^7$ State Key Laboratory of General Artificial Intelligence, Peking University\quad{}\\
\vspace{-20pt}
}

\twocolumn[{
\renewcommand\twocolumn[1][]{#1}
\maketitle
\begin{center}
    \captionsetup{type=figure}
    \includegraphics[width=0.99\linewidth]{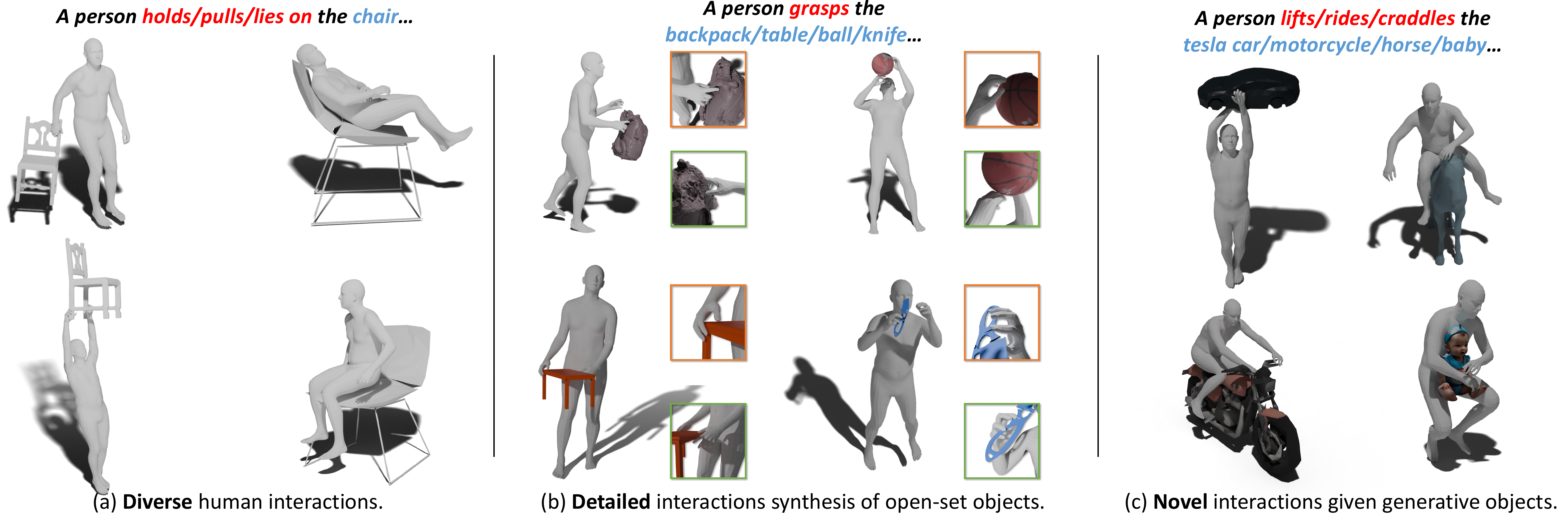}
    \captionof{figure}{\textbf{3D human object interaction synthesis by \method.} Given a simple text description with goal interaction and any object mesh as input, our method enables the generation of diverse, natural, detailed, and novel interactions for open-set 3D objects in a zero-shot manner. The \textcolor{orange}{orange} and \textcolor{green}{green} boxes of (b) indicate detailed contact poses from different views.}
    \label{fig:teaser}
\end{center}
}]

\begin{abstract}
Recent advances in 3D human-aware generation have made significant progress. However, existing methods still struggle with generating novel Human Object Interaction (HOI) from text, particularly for open-set objects. We identify three main challenges of this task: precise human-object relation reasoning, affordance parsing for any object, and detailed human interaction pose synthesis aligning description and object geometry. In this work, we propose a novel zero-shot 3D HOI generation framework without training on specific datasets, leveraging the knowledge from large-scale pre-trained models. Specifically, the human-object relations are inferred from large language models (LLMs) to initialize object properties and guide the optimization process. Then we utilize a pre-trained 2D image diffusion model to parse unseen objects and extract contact points, avoiding the limitations imposed by existing 3D asset knowledge. The initial human pose is generated by sampling multiple hypotheses through multi-view SDS based on the input text and object geometry. Finally, we introduce a detailed optimization to generate fine-grained, precise, and natural interaction, enforcing realistic 3D contact between the 3D object and the involved body parts, including hands in grasping. This is achieved by distilling human-level feedback from LLMs to capture detailed human-object relations from the text instruction. Extensive experiments validate the effectiveness of our approach compared to prior works, particularly in terms of the fine-grained nature of interactions and the ability to handle open-set 3D objects. Project page: \href{https://jinluzhang.site/projects/interactanything}{jinluzhang.site/projects/interactanything}.
\end{abstract}

\section{Introduction}
\label{sec:intro}
Synthesizing the interaction of humans and any given object holds great significance across diverse applications, including AR/VR, computer simulation, and animation. Traditionally, creating high-quality 3D human object interaction (HOI) assets requires expensive, lab-controlled data capture systems and significant human effort. Although some datasets~\cite{BEHAVEDatasetMethod2022Bhatnagar,scalingdataset2024sjiang,omomo2023,chairsdataset,grabdataset2020,hassan2019resolving} have collected these data, the models~\cite{HOIDiffTextDrivenSynthesis2023Peng,xu2023interdiff,omomo2023,ControllableHumanObjectInteraction2023Li,DECODenseEstimation2023Tripathi,GeneratingHumanInteraction2024Yi} trained on them can only generate interactions for specific object categories. These methods struggle to handle the diversity of object geometries, scales, and interactions at the same time, making text-driven zero-shot HOI synthesis a challenging and highly demanding task.

Given recent advances in text-to-3D techniques~\cite{magic3d,fantasia3D,prolificDreamer}, \eg, leveraging Score Distillation Sampling (SDS)~\cite{poole2022dreamfusion} to generate 3D content from text, a natural question is whether this paradigm can be applied to zero-shot HOI synthesis.
However, directly integrating 3D humans and objects into a scene without understanding the complex relationships between them remains challenging. Recent works~\cite{zhu2024dreamhoi,dai2024interfusion,chen2024comboverse} are proposed to generate realistic 3D HOIs. Among them, InterFusion~\cite{dai2024interfusion} uses Neural Radiance Fields (NeRF)~\cite{mildenhall2020nerf} for individual human and object representations but struggles with precise interaction area, leading to suboptimal and unrealistic HOIs. DreamHOI~\cite{zhu2024dreamhoi} optimizes on both mesh and NeRF representations through bidirectional rendering but faces challenges in ensuring accurate contact points and relative scales due to solely optimizing human states while keeping the given object static, leading to limited interactions.

In this paper, we first identify the core challenge of zero-shot 3D HOI generation as accurately capturing human-object relations, contact areas, and global consistency, which are not typically presented in single-object or non-contact generation scenarios. These challenges prevent the generation of high-quality, geometrically precise, and textually consistent results. Furthermore, while large-scale pre-trained models, \eg, 2D diffusion models~\cite{podell2023sdxlimprovinglatentdiffusion, stablediffusion_2022_Rombach} and large language models~(LLMs)~\cite{google2023gemini,achiam2023gpt4,openai2023gpt35}, cannot fully capture the complexity of human object interactions (including body dynamics, expressive details, and object geometry), they can offer valuable prior knowledge of common interaction patterns based on textual descriptions. Consequently, we argue that the solution for effective generation is to leverage this diverse prior knowledge to guide the optimization of interactions.

Following the above insights, we decompose the complex zero-shot 3D HOI task into these stages: 1) human-object relation reasoning and initialization, 2) open-set object contact affordance parsing, and 3) human pose synthesis aligning text and object geometry. 
We first leverage the LLM as a human-level feedback provider, which reasons and unravels complex semantic interaction relationships, \eg, relative position and orientation, from simple text input. It initializes human and object states and designs a series of optimization functions to generate detailed interactions.
The initial human pose is optimized by multi-view SDS loss for interaction in line with the given textual description and object geometry.

To parse the 3D object affordance, we utilize initialized human-object relation and sample adaptive inpainting masks during the diffusion process to obtain the contact information. This enables the extraction of contact probabilities on the object, which are then used to segment and define the weighted dynamic contact regions. 
Finally, we design a coarse-to-fine optimization based on force closure~\cite{force_closure,dfc} to generate expressive, diverse, and physically plausible interactions, allowing for more accurate and detailed contact.
To summarize, our contributions are:
\begin{itemize}
\item We propose \method for zero-shot 3D HOI generation, which effectively distills human and object interactions from LLM and pre-trained diffusion models to synthesize diverse, detailed, and novel interactions.

\item We design a novel open-set object affordance parsing, which generates contact maps on 3D geometry by utilizing adaptive inpainting and potential contact distribution.

\item Our model leverages both text and object geometry to drive human pose synthesis, enabling realistic and contextually appropriate interactions for open-set objects. Results show that our synthesized pose accurately reflects the semantic meaning of the input while respecting the spatial constraints of the object, leading to improved performance, as shown in Figure~\ref{fig:teaser}.
\end{itemize}

\section{Related Work}
\label{sec:relatedwork}
\subsection{Text-driven 3D Content Generation}
The generation of 3D content from text has seen rapid advancements, primarily driven by diffusion-based models that build on text-to-2D image generation approaches~\cite{rombach2022highresolutiondiffusion,ruiz2023dreambooth,Photorealisticdiffusion,Ramesh_Dhariwal_Nichol_Chu_Chen}. One pioneering approach, DreamFusion~\cite{poole2022dreamfusion}, introduced Score Distillation Sampling (SDS), which leverages the pre-trained 2D diffusion model to optimize 3D content through techniques like NeRF (Neural Radiance Fields)~\cite{mildenhall2020nerf} and 3D Gaussian Splatting~\cite{kerbl20233d}. 
A significant area of research aims to address the complexity of 3D human generation. For example, recent methods have been proposed for generating controllable 3D human avatar~\cite{hong2022avatarclip,cao2024dreamavatar,liu2024humangaussian,hu2024gauhuman,Humannerf2022Zhao,HumanNeRF2022Weng}. 
However, these works focus on diverse and high-quality appearance but animate human avatars with given human poses or motion, which do not satisfy 3D HOI requirements in human pose synthesis.
The proliferation of large 3D datasets has also catalyzed progress, leading to more sophisticated models like Zero-1-to-3~\cite{liu2023zero}, Zero123++~\cite{shi2023zero123++}, and MVDream~\cite{shi2023mvdream}, which employ 2D diffusion models to generate multi-view consistent images that serve as inputs for efficient 3D generation systems, such as SyncDreamer~\cite{liu2023syncdreamer} and Wonder3D~\cite{long2024wonder3d}.
Despite these advancements, generating complex HOI scenes, especially those involving interactions between humans and objects, remains challenging. 

\subsection{Data-Driven Human Object Interaction}
Previous approaches~\cite{omomo2023,GeneratingHumanInteraction2024Yi,xu2023interdiff,xu2024interdreamer} primarily train generative models to produce static human poses or dynamic motions in response to objects or scenes using curated HOI datasets. Recent works focus on dynamic motion generation performance conditioned on encoded object features but rely heavily on specialized datasets, limiting generalization to diverse objects. For example, OMOMO~\cite{omomo2023} generates human motion through diffusion models but requires object motion as input. 
InterDiff~\cite{xu2023interdiff} predicts full-body interactions based on prior HOI context but is limited to the prediction task, restricting flexibility in new scenarios. 
In contrast, our method jointly synthesizes human and object states and adapts to novel objects without extensive, object-specific training by extracting interaction patterns directly from 2D diffusion models.

On the other hand, optimization-based methods reconstruct realistic human poses and transformations from images. PHOSA~\cite{zhang2020phosa} optimizes human object relations using depth-aware loss functions. PROX~\cite{hassan2019resolving} encodes contact probabilities and semantic scene labels to position humans within scenes. CHORUS~\cite{han2023chorus} synthesizes multi-view images with a pre-trained generative model to learn 3D spatial relationships. By extracting knowledge from LLM to infer human poses, the method in \cite{wang2022reconstructing} queries contact information to reconstruct 3D human-object interactions. Compared to these methods, our method leverages pre-trained diffusion models for accurate parsing of object affordance and enables fine-grained interactions.

\subsection{Zero-shot 3D Compositional Generation}
To achieve the compositional scene generation, ComboVerse~\cite{chen2024comboverse} proposes a spatially-aware SDS loss, generating the final 3D scene through single-object reconstruction and multi-object composition.
GraphDreamer~\cite{gao2024graphdreamer} leverages scene graphs to guide generation by converting the semantic information into a global textual description to enable text-driven compositional scene generation.
These methods focus on the reasonable arrangement of multiple objects but ignore the contact generation in HOI scenes.
InterFusion~\cite{dai2024interfusion} and DreamHOI~\cite{zhu2024dreamhoi} present this challenge and then try to address using SMPLs~\cite{loper2015smpl,MANO2017} as body model prior.
DreamHOI~\cite{zhu2024dreamhoi} designs the SDS guidance mixture to separately render human interactions using different diffusion models to improve human modeling quality. However, it cannot ensure precise contact during the denoising process.
Compared to these methods, our work focuses on addressing contact areas parsing on open-set objects.

\begin{figure*}
    \centering
    \includegraphics[width=0.99\linewidth]{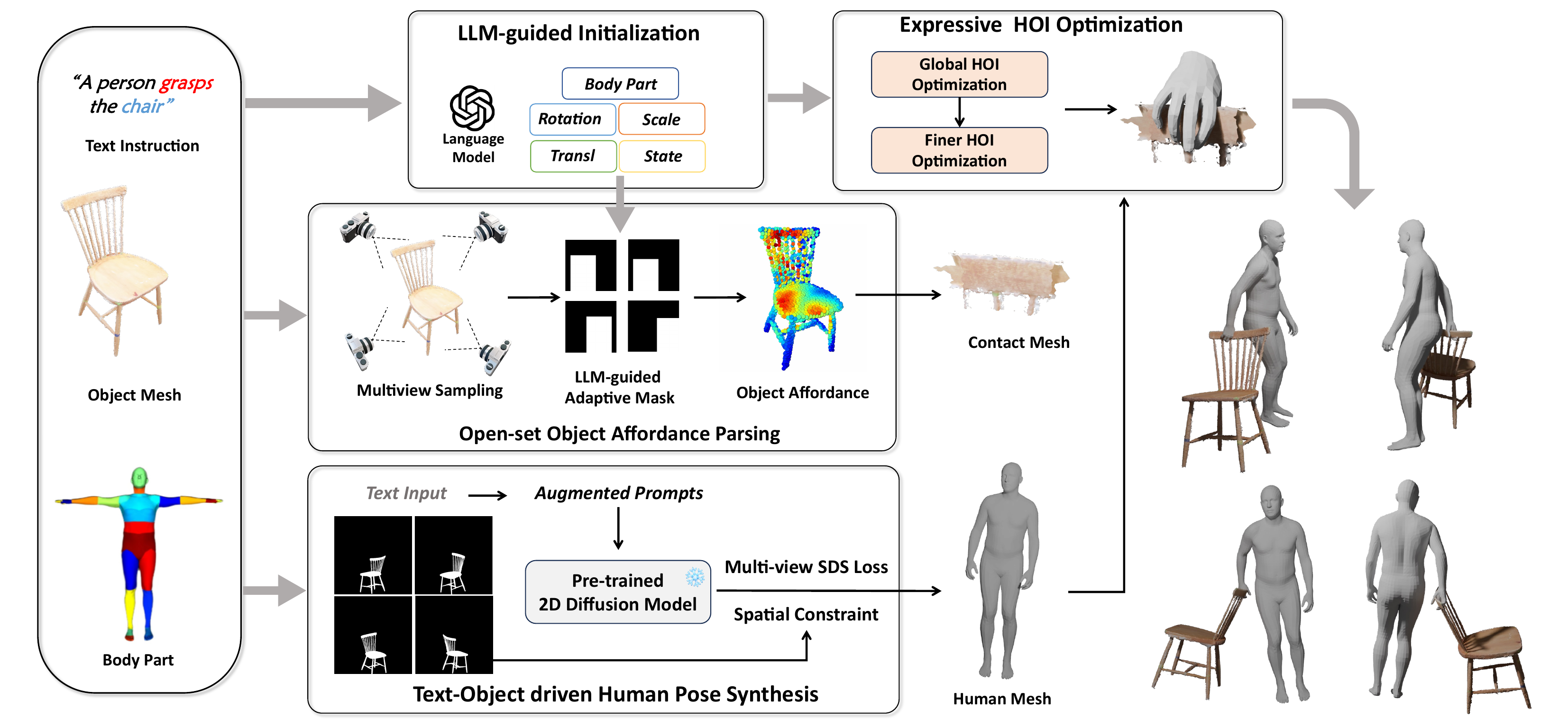}
    \caption{\textbf{Framework of \method.} Given a text description and any object mesh as input, our approach begins by querying LLM to infer precise human-object relationships, which are used to initialize object properties. Next, we analyze the contact affordance of the object geometry. The human pose is synthesized using a pre-trained 2D diffusion model, guided by multi-view SDS loss and the designed spatial constraint. Finally, based on the targeted object contact areas and a plausible human pose, we perform expressive HOI optimization to synthesize realistic and contact-accurate 3D human-object interactions.}
    \label{fig:pipeline}
\end{figure*}
  
\section{Our Method}
\textbf{Task Definition and Overview.}
Given a text prompt $\epsilon$ that describes the desired HOI, \eg, ``A person grasps the chair". Our goal is to synthesize natural human mesh $\mathcal{V}_H$ and object mesh $\mathcal{V}_O$ that align with $a$ and $\epsilon$. 
As shown in \Cref{fig:pipeline}, we first obtain initial human-object relations from LLM based on text (\Cref{sec:llminit}). For object optimization, we initialize object rotation $r_o$, translation $t_o$, scale $s_o$, state $\mathbf{s}$ , based on affordance map $\mathcal{M}$ (\Cref{sec:affordance}). We also synthesize the initial human pose based on the input text and object geometry (\Cref{sec:pose}). Finally, we use SMPL-H as representation and optimize pose parameter $\theta_b, \theta_h$, global translation $t_b$, and object rotation and translation $r_o, t_o$ to generate detailed and diverse HOI (\Cref{sec:hoiopt}).

\subsection{Preliminary Knowledge}
\label{sec:preliminaries}
\noindent \textbf{Human Body Model.} We use the SMPL-H~\cite{loper2015smpl,MANO2017}, a parametric model that includes both body and hand parameters, to represent the human body and hand articulation.
The body pose is defined by body parameters $\mathbf{\theta}_{\text{b}} \in \mathbb{R}^{3(K_b+1)}$, and the hand pose is represented by additional MANO parameters $\mathbf{\theta}_{\text{h}} \in \mathbb{R}^{3K_h}$, capturing 15 joints per hand.
The shape parameters are represented by $\mathbf{\beta} \in \mathbb{R}^{10}$, controlling the deformation of the body and hand template mesh to target body shapes. Given the combined pose parameters $\{\mathbf{\theta}_{\text{b}}, \mathbf{\theta}_{\text{h}}\}$ and shape parameters $\mathbf{\beta}$, SMPL-H computes mesh vertices $\mathcal{V} \in \mathbb{R}^{V \times 3}$ using linear blend skinning function $f$:
\begin{equation}
\mathcal{V}(\beta, \theta_b, \theta_{\text{h}}) = f \left( T_p(\beta, \theta_b, \theta_{\text{h}}), J(\beta), \theta, \mathcal{W} \right),
\end{equation}
where $J$ represents joint locations, $\mathcal{W} \in \mathbb{R}^{V \times (K)}$ is skinning weights with all $K$ joints.

\noindent \textbf{Score Distillation Sampling (SDS).} SDS leverages pre-trained diffusion models to guide the synthesis of 3D scenes by aligning NeRF with a text prompt. It operates by iteratively adjusting NeRF parameters, denoted as \( \theta \), based on gradients computed to align the generated images with the textual input. 
Given a noisy 2D image \( x_t \) (obtained by adding noise \( \epsilon \) to a clean NeRF-rendered image \( x \)), a textual prompt \( y \), and a noise level \( t \), SDS calculates the gradient of the loss \( \mathcal{L}_{\text{SDS}} \) with respect to \( \theta \) as follows:
\[
\nabla_\theta \mathcal{L}_{\text{SDS}}(x) = \mathbb{E}_{t, \epsilon} \left[ w(t) \left( \hat{\epsilon}(x_t; y, t) - \epsilon \right) \frac{\partial x}{\partial \theta} \right],
\]
where \( \hat{\epsilon}(x_t; y, t) \) represents the diffusion model predicted noise given the noisy image \( x_t \), prompt \( y \), and noise level \( t \), \( w(t) \) is a weighting function that adjusts the influence of the gradient based on the noise level \( t \).
This gradient is used to update \( \theta \), guiding the NeRF model toward producing images that match the textual description. For instance, in DreamFusion~\cite{poole2022dreamfusion}, this process enables text-to-3D generation by repeatedly refining the NeRF parameters \( \theta \) from a random initialization, iteratively producing images that align more closely with the target prompt.

\subsection{LLM-guided Initialization}
\label{sec:llminit}
To infer diverse intended human object interactions for a given object $\mathcal{V}_O$ and action $a$, we query the LLM to provide commonsense feedback. To minimize hallucination, we need to transfer inferred relationships from GPT~\cite{openai2023gpt35} into specific values that position the human and object meshes accurately, therefore, we enable the language model to select answers from pre-defined options step by step, instead of directly generating responses. 
For initializing the object rotation $r_o$, translation $t_o$, scale $s_o$, and state $\mathbf{s}$, the language model selects the relative position of the human to the object from defined options, which is then mapped to $t_o$ based on the combined size of the human and object. We set all input objects to have $y$-up and $x$-front orientations, with $r_o$ adjusted based on which side of the object should face the human.
The object scale is chosen from 4 defined options and set relative to human size. Lastly, the state $\mathbf{s}$ determines if the object should be positioned on the ground based on whether it is meant to move. Besides, we also query interacted semantic body parts $\mathcal{S}_h$ and map the labels to human mesh to obtain vertices $\mathcal{P}_h$ for part-level optimization using annotated human part labels.

\begin{figure*}[t]
\centering
\includegraphics[width=0.99\linewidth]{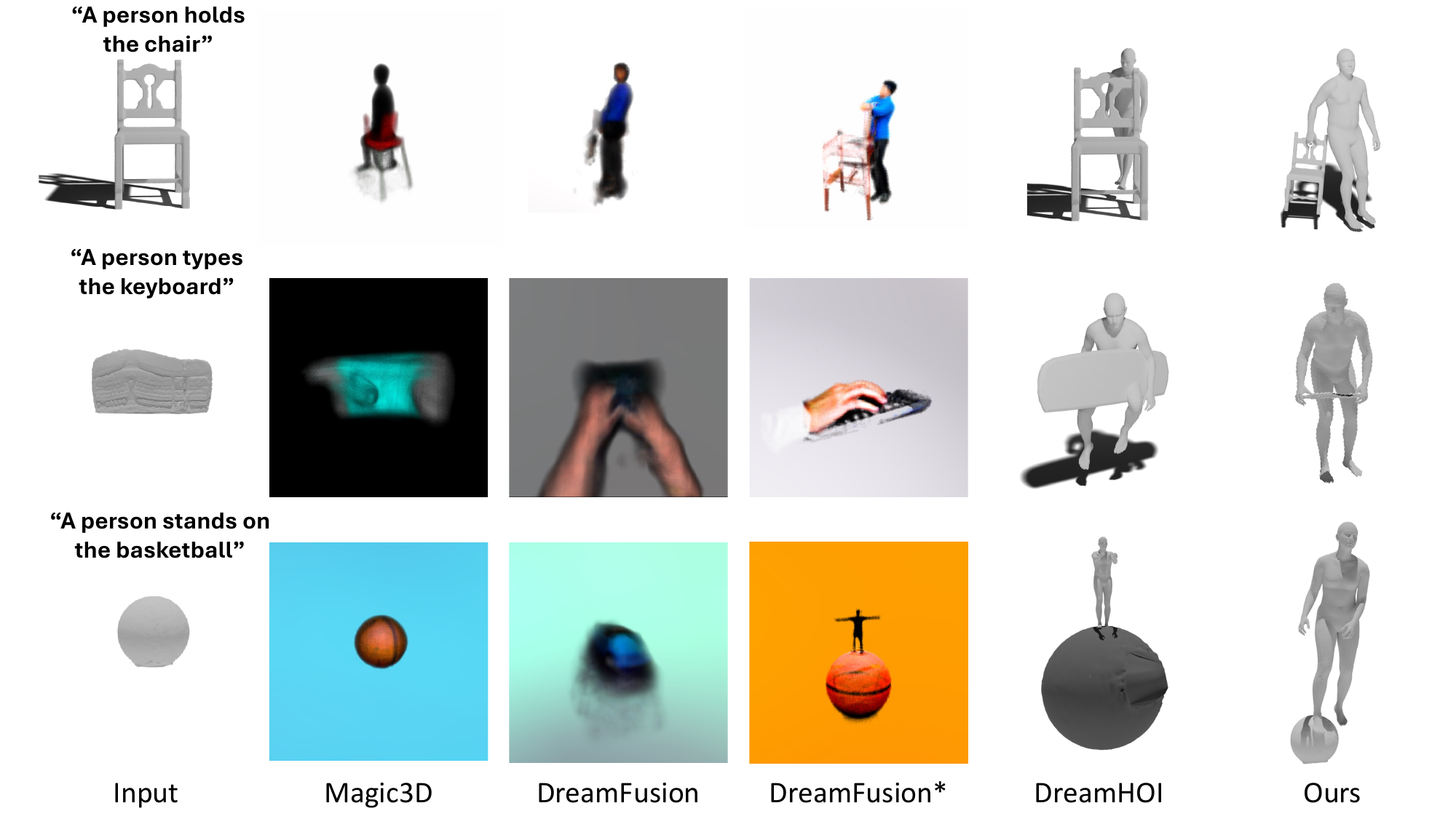}
\caption{\textbf{Qualitative comparison results with baselines.} $*$ indicates we re-implement this method by embedding object mesh into the diffusion process, which follows the pipeline of DreamHOI~\cite{zhu2024dreamhoi} and our method. More visualization results are presented in supplementary materials.} 
\label{fig:comparison}
\end{figure*}

\subsection{Open-set Object Affordance Parsing}
\label{sec:affordance}
To obtain the contact affordance of unseen object geometries given a specified action \eg, \textit{holds}, we conduct 2D inpainting based on a pre-trained diffusion model to distill possible human interaction parts through LLM-guided adaptive mask generation. Then we design a 2D probabilistic contact function to model the distribution of the interaction, generating an affordance map to identify and extract the regions for HOI optimization.

\noindent \textbf{2D Adaptive Mask Generation}
Image inpainting is to fill in missing regions of an image, where the missing regions can be defined by a mask image. To get knowledge of possible interacted parts, we utilize the powerful understanding and generation ability of pre-trained inpainting model~\cite{podell2023sdxlimprovinglatentdiffusion}.
Given 2D projection images of the 3D object from $n_p$ views, we found existing inpainting models cannot ensure complete human in the image (many inpainting results only include interacted body parts, \eg, hands in ``hold'' action). Therefore, different from previous methods~\cite{coma,li2024genzi}, we do not randomly sample the inpainting masks on 2D images, instead, our method identifies possible human interaction parts to improve the inpainting success rate by generating inpainting masks from LLM feedback in two forms: (1) \textit{full-body} mask with respect to object properties $r_o$, $t_o$, person position and person body size; (2) \textit{body-part} inpainting masks for intended interacted body part labels $\mathcal{S}_h$ and corresponding human vertices $\mathcal{P}_h$.

More specifically, we first create a full-body inpainting mask guided by LLM-inferred initializations \( r_o \), \( t_o \), and body size $s_f$:
\begin{equation}
  M_{\text{full}}^{i} = \mathcal{J}((\mathcal{V}_H \cdot r_o^{-1}), s_f, c^i) + t_o^{-1} + \varpi_1,
\end{equation}
where $\mathcal{J}$ is 2D projection transformation with camera parameter $c^i$, and $\varpi_1$ is the noise term.
Similarly, we generate body-part-level inpainting masks using given body part labels \( \mathcal{S}_h \) and corresponding human vertices \( \mathcal{P}_h \) and body part size $s_p$ to help inpainting body parts like hands as the above formulation:
\begin{equation}
  M_{\text{part}}^{i} = \mathcal{J}((\mathcal{P}_h \cdot r_o^{-1}), s_p, c^i) + t_o^{-1} + \varpi_2.
\end{equation}
Besides, We apply the above full-body masks $M_{\text{full}}$ into the dynamic mask inpainting pipeline~\cite{coma}, which can further refine the initial mask detecting human on intermediate results in the diffusion process.
    
\noindent \textbf{Affordance Parsing on 2D Probability Function.}
Directly computing contact probabilities on a 3D object mesh relies heavily on existing 3D human reconstruction methods~\cite{KeepItSMPL2016Bogo,yin2024whac,PyMAFX2023Zhang,kanazawa2018end,lin2023one,MixSTE2022Zhang,gong2023diffpose} from single-view images. This encounters significant 3D ambiguity and quantization error, which obscures accurate 3D human pose. In contrast to previous methods~\cite{han2023chorus}, we apply 2D human pose detector OpenPose~\cite{openpose}, performing well at detecting 2D human keypoints and even occluded body parts, to estimate human from inpainting results. Then we introduce a 2D probability function that calculates contact probability according to the distance from detected body keypoints to the multi-view object projection images.
First, we downsample the object to $D$ points. Let \(\mathcal{P}_i\) represent the probability map for each view \(i\), the distance-based probability \( f_{\text{afford}}(\mathbf{p}) \) for a point \(\mathbf{p} \in \mathcal{V}_O\) on the object surface is defined as:
\begin{equation}
  f_{\text{afford}}^{(i)}(\mathbf{p}) = e^{-\| \mathbf{d}_i(\mathbf{p}) \|},
\end{equation}
where \(\mathbf{d}_i(\mathbf{p})\) denotes the distance from the body part to the object mask in \(i\) view.
Then we re-project the 2D affordance $f_{\text{afford}}^{(i)}(\mathbf{p})$ to 3D space using defined camera parameters, and aggregate the contact probabilities across all \(n_p\) views to update the 3D contact probability distribution as follows:
\begin{equation}
  \mathcal{P}(\mathbf{p}) = \frac{1}{n_p} \sum_{i=1}^{n_p} f_{\text{afford}}^{(i)}(\mathbf{p}).
\end{equation}
Finally, we normalize \(\mathcal{P}(\mathbf{p})\) over the entire object surface to obtain contact probability for each point is determined by its projected 2D distances across multiple views, effectively mitigating multi-view ambiguity and enabling accurate estimation of human object contact regions.

\subsection{Text-Object Driven Human Pose Synthesis}
\label{sec:pose}
Considering a brief HOI description, we posit that current 3D generation models, such as DreamFusion~\cite{poole2022dreamfusion}, are capable of generating human avatars in a variety of poses; however, they do not guarantee precise contact with the object mesh. 
Thus, we first apply prompt engineering to generate diverse and expressive text descriptions, along with negative prompts, using large language models (LLMs) to mitigate issues like low-quality outputs and missing limbs.
Next, we incorporate Multi-view SDS loss~\cite{shi2023mvdream} to optimize the human pose aligning the interaction description and object geometry. We first scale the input object based on $s_o$ given by the language model. 
We use the spatial constraint to prevent the human avatar from penetrating the object mesh, ensuring alignment with spatial relations. Specifically, we embed object mesh within NeRF optimization by regularizing the ray sample points in object mesh following previous works~\cite{zhu2024dreamhoi,dai2024interfusion}. After training, we use the multi-view human pose estimation method~\cite{zheng2021pamir} to obtain the body parameters $\theta_b$ and then init hand parameters $\theta_h$. We also employ VPoser~\cite{SMPL-X} to maintain reasonable joint rotations.

\subsection{Expressive HOI Optimization}
\label{sec:hoiopt}
\noindent \textbf{Global HOI optimization.}
With the above initialization as input, we first ensure the input object is watertight using Manifold~\cite{manifold}. Then we conduct human-centric optimization to update object properties ($r_o, t_o, s_o$) as follows:
\begin{equation}
  L_{c} = \phi_i L_{inter} + \phi_n L_{n} +\phi_s L_{scale} + \phi_p L_{pene} + \mathbf{s}^{'} \cdot L_{g},
\end{equation}
where $\phi$ is loss weights and interaction loss is designed on verts of the interacted body part $\mathcal{P}_h, h \in \mathcal{V}_H$ and object contact areas $\mathcal{P}_o, o \in \mathcal{M}$:
\begin{equation}
  L_{inter} = \sum_{i \in [h], j \in [o]} W_{\mathcal{M}}(j) \cdot {f}_{\textit{cham}}(\mathcal{P}_h, \mathcal{P}_o),
\end{equation}
where $W_{\mathcal{M}}(j)$ indicates probabilities scores in object affordance map $\mathcal{M}$ of the $j$-th object vertex and ${f}_{\textit{cham}}$ is chamfer distance function. We constrain the normals of interacted parts to maintain the relative orientation:
\[
L_{n} = \sum_{n \in [h, o]} ( \mathbf{n}_h \cdot \mathbf{n}_o + 1 )^2,
\]
where $\mathbf{n}_h$ is normals of interacted body verts, and $\mathbf{n}_o$ indicates object normals of interacted verts. 
In addition, we employ the object scale loss $L_{\text{scale}}$~\cite{zhang2020phosa} and the penetration term $L_{\text{pene}}$ to prevent collisions~\cite{Karras2012intersect}. The introduction of ground loss $L_{g}$ is decided by the object state $\mathbf{s}=[0, 1]$.

\noindent \textbf{Finer HOI Optimization.}
During this stage, we optimize the human parameters $\theta_b$, $\theta_h$, and $t_b$ and fix the object parameters.
To synthesize detailed and diverse interaction poses for full-body optimization, inspired by differentiable force closure~\cite{dfc}, we design the fine-grained optimization loss as below:
\begin{equation}
  L_{f} = \delta \cdot L_{fc} + L_{pene} + L_{inter},
\end{equation}
where loss functions $L_{\text{pene}}$ and $L_{inter}$ are analogous to the aforementioned terms, but with the incorporation of Kaolin~\cite{KaolinLibrary} to better sample surface distance.
We introduce SMPL-based force closure loss $L_{fc}$ by replacing the base model from MANO to SMPL-H to conduct more detailed interaction poses:
\begin{equation}
  L_{fc} = \sum_{j \in [o]} \left( \sum_{i \in [h]}fv(i, j) \cdot n(j) \right)^2,  \delta \cdot [\theta_b, \theta_h].
\end{equation}
$fv$ symbolizes the force vector applied at vertex $j$ from interacted human vertex $i$, and $\delta$ decide optimize body pose $\theta_b$ or hand pose $\theta_h$.
Specifically, we make a series of augmentations to improve force closure acceptance rate and robustness. We decrease contact points by sparse uniform sampling on the interacted vertices. Moreover, we initialize body and hand positions around the interacted object parts and maintain the hand normals as positive to the interacted object parts. Finally, we extract the interacted vertices to form interacted mesh and non-interacted mesh and compute $L_{fc}$ and $L_{pene}$ separately.

\section{Experiments}
\label{sec:exp}
\subsection{Data and Metrics}
We collect evaluated object meshes in 3 ways: BEHAVE~\cite{BEHAVEDatasetMethod2022Bhatnagar}, PartNet~\cite{mo2019partnet}, and sampling from generative models. 
BEHAVE dataset is commonly used in this field, and PartNet is manually annotated in an object mesh with part-level semantic labels on the vertices.
For metrics, we use CLIP similarity score~\cite{clip} between our prompts and the rendering images of the HOI scene, by rendering our human and object meshes from 4 different views randomly on 5 prompts and 10 objects categories. During this process, we select 5 BEHAVE objects (backpack, basketball, chairwood, keyboard, and suitcase), and 5 objects from publicly accessed 3D generative models (motorcycle, car, television, baby doll, and humanoid robot).
The same evaluation setting is adapted to other comparison methods. 
In addition, while the CLIP score is designed to evaluate how well an image aligns with the input text, it often fails to capture finer nuances in interactions, leading to less distinct variations in the metrics.
We follow InterFusion~\cite{dai2024interfusion} to leverage the advanced image understanding capabilities of GPT-4V further to assess the performance of baselines and our method. The text prompts are designed and then augmented using GPT.

\begin{table}[t]
    \centering
    \setlength{\tabcolsep}{3pt}
    \resizebox{\linewidth}{!}{%
        \begin{tabular}{lcccc}%
            \toprule 
            \textbf{Metric} & \textbf{Magic3D~\cite{magic3d}} & \textbf{DreamFusion~\cite{poole2022dreamfusion}} & \textbf{DreamHOI~\cite{zhu2024dreamhoi}} & \textbf{Ours} \\
            \midrule
            CLIP Score$\uparrow$ & 0.2416 & 0.2437 & 0.2757 & \textbf{0.2968} \\
            \bottomrule
        \end{tabular}%
    }
    \caption{\textbf{Comparison of CLIP similarity scores for different methods.} Higher scores are better.}
    \label{tab:clip_scores}
\end{table}

\begin{figure}[t]
    \includegraphics[width=0.99\linewidth]{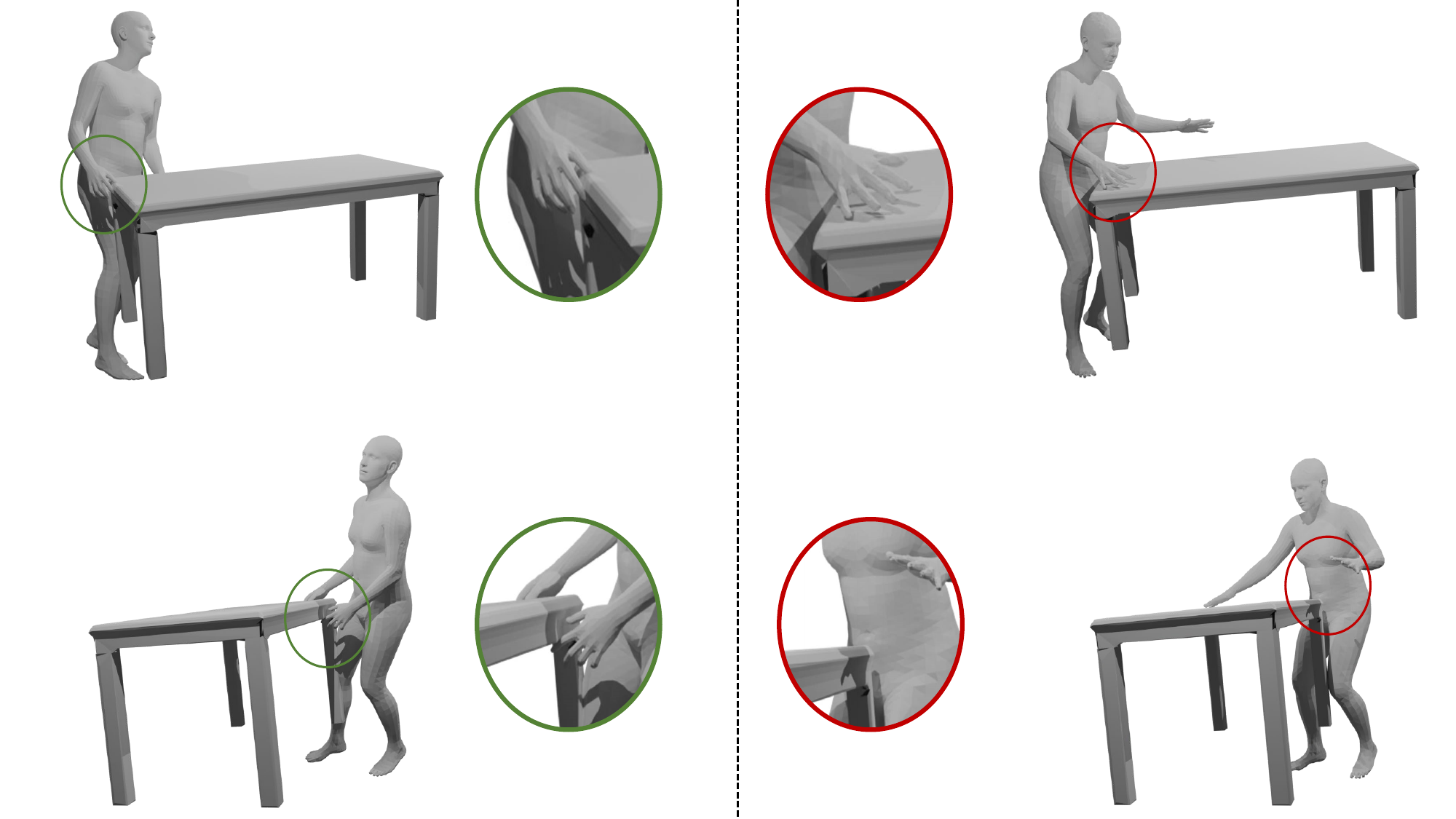}
    \caption{\textbf{Ablation study on the fine-grained optimization.} Left figures are the results of removing fine-grained terms and right figures apple fine-grained terms to synthesize grasping details.}
    \label{fig:ablation-1}
\end{figure}

\subsection{Quantitative Evaluations}
\noindent\textbf{CLIP Score.}
As shown in \Cref{tab:clip_scores}, we present a comparison of CLIP similarity scores across various methods, where higher scores indicate better alignment with the target descriptions. Our method \method, achieves the highest mean score across the evaluation prompts, outperforming baselines such as DreamFusion~\cite{poole2022dreamfusion}, Magic3D~\cite{magic3d}, and DreamHOI~\cite{zhu2024dreamhoi}. These results demonstrate that our approach effectively captures the semantic interactions of the text descriptions. Notably, to better evaluate interaction performance without being confounded by appearance, we apply random human textures to the body mesh and retain the original textures for objects.

\noindent\textbf{GPT-4V Selection.}
As presented in \Cref{tab:comp_gpt}, we enable the vision language model~\cite{achiam2023gpt4} to choose the most reasonable result in terms of \textit{Overall} and \textit{Contact} from all generated outputs based on criteria such as the presence of a full human body, a complete object, and correct physical interactions, and then returning the index of the selected result. 
No in-context examples are provided to guide the model in the \textit{Overall} setting, we give text prompts only for the \textit{Contact} evaluation, and the order of generated results is randomly shuffled to avoid bias, the final results are obtained from multiple tests. 

\begin{table}[t!]
    \centering
    \small
    \setlength{\tabcolsep}{3pt}
    \resizebox{\linewidth}{!}{%
        \begin{tabular}{cccccc}%
            \toprule
            \textbf{Metrics} & \textbf{Magic3D} & \textbf{DreamFusion} & \textbf{DreamFusion*} & \textbf{DreamHOI} & \textbf{Ours} \\
            \midrule
            Overall  & 4.3   & 6.5  & 17.3  & 26.0  & \textbf{45.6}    \\
            Contact  & 8.6	 & 4.3	& 15.2	& 19.5	& \textbf{52.1}    \\
            \bottomrule
        \end{tabular}%
    }%
    \caption{\textbf{Quantitative evaluation of GPT-4V selection in terms of \textit{Overall} and \textit{Contact}.}}
    \label{tab:comp_gpt}
\end{table}

\begin{figure}[t]
    \centering
    \includegraphics[width=0.99\linewidth]{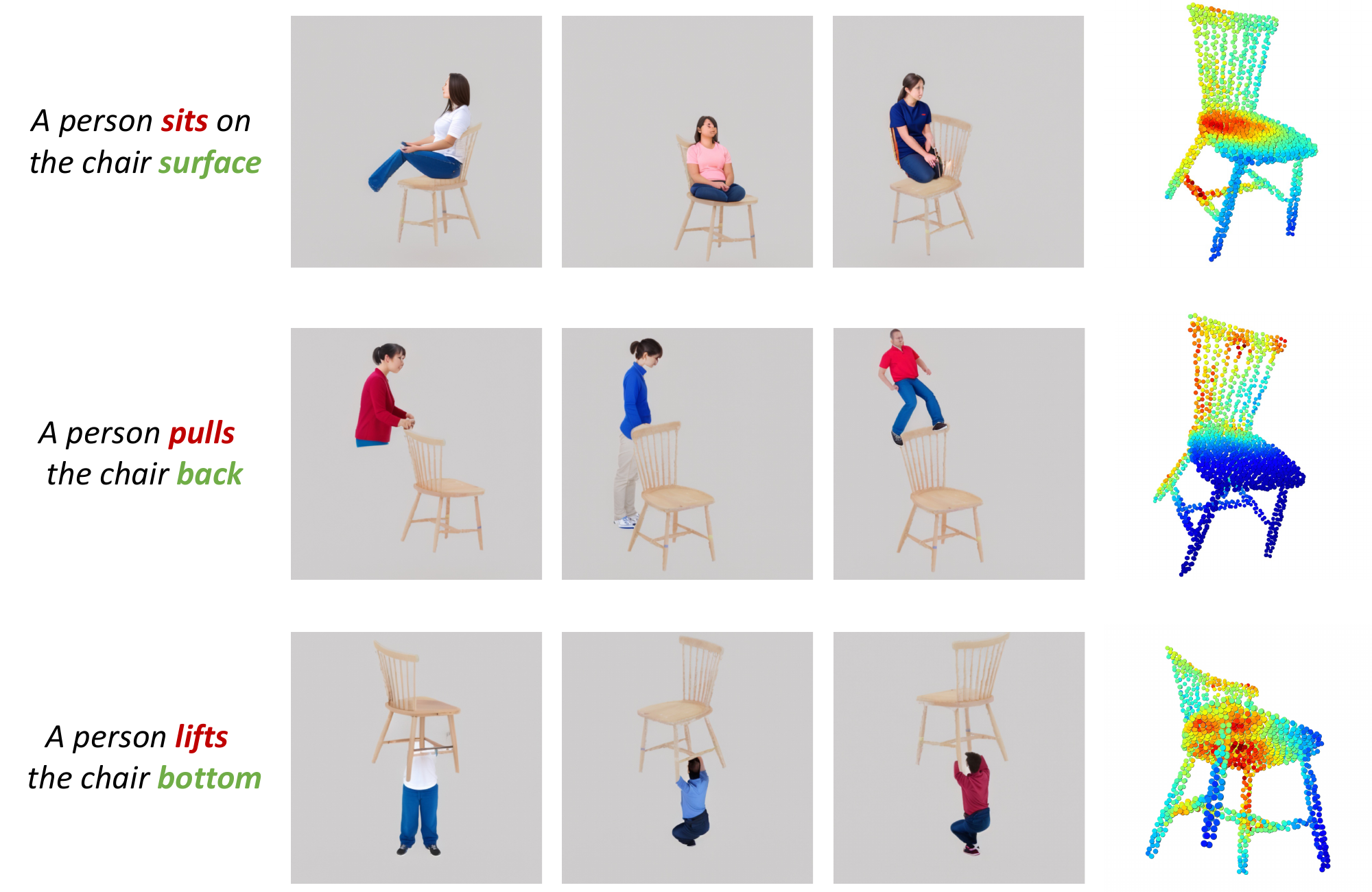}
    \caption{\textbf{Visualization of the adaptive inpainting and open-set object affordance parsing results on the same object with different text instructions.} Our method first generates reasonable 2D inpainting results (middle columns) and then computes contact probabilities as the affordance representation (right column).}
    \label{fig:vis-affordance-chair}
\end{figure}

\subsection{Qualitative Evaluations and Application}
\noindent\textbf{Visualization Comparison.}
The comparison results are shown in \Cref{fig:comparison}, where our method outperforms the baselines, producing the most realistic and plausible 3D human-object interactions. This is evident in both the level of detail and the physical accuracy of the interactions.
Magic3D~\cite{magic3d}, for instance, often generates overly simplified shapes with minimal texture and detail, as seen in the keyboard typing example, where the lack of intricacy undermines realism. Both Magic3D~\cite{magic3d} and DreamFusion~\cite{poole2022dreamfusion} fail to capture body structure or realistic human-object relationships, highlighting the difficulty of generating realistic HOI using general 3D scene generation models.
DreamFusion*, which incorporates object mesh as input, retains a complete or partial human body shape but struggles with accurate contact and natural scale. This limitation is particularly noticeable in examples like the holding-chair scenario, where hand placement lacks precision. DreamHOI~\cite{zhu2024dreamhoi} improves upon DreamFusion in terms of HOI but still falls short in capturing fine interaction details, such as accurate foot placement on the basketball.

\begin{table}[t!]
    \centering
    \small
    \setlength{\tabcolsep}{3pt}
    \resizebox{\linewidth}{!}{%
        \begin{tabular}{cccccc}%
            \toprule
            \textbf{Setting} & \textbf{Magic3D} & \textbf{DreamFusion} & \textbf{DreamFusion*} & \textbf{DreamHOI} & \textbf{Ours} \\
            \midrule
            {Full}          & 4.3   & 6.5  & 17.3  & 26.0  & {45.6}  \\
            {w/o Refine}    & 6.5	& 6.5	& 17.3	& 30.4	& 39.1  \\
            {w/o LLM-init}  & 8.6	& 10.8	& 21.7	& 32.6	& 26.1  \\
            \bottomrule
        \end{tabular}%
    }%
    \caption{\textbf{Ablation study on LLM-guided initialization and HOI refinement by GPT-4V selection.}}
    \label{tab:ablation_llm}
\end{table}

\noindent\textbf{Inpainting Visualization.}
We further visualize the 2D inpainting results and affordance parsing on the same object in \Cref{fig:vis-affordance-chair} to emphasize the effectiveness of our method in disentangling affordances for distinct human-object interactions. The figure showcases various human object interactions with a chair, highlighting the ability to localize affordances accurately.
This visualization provides valuable insight into the regions on the object that are most likely to be in contact with humans, which is critical for optimizing human-object interaction.

\begin{figure}[t!]
    \centering
    \includegraphics[width=\linewidth]{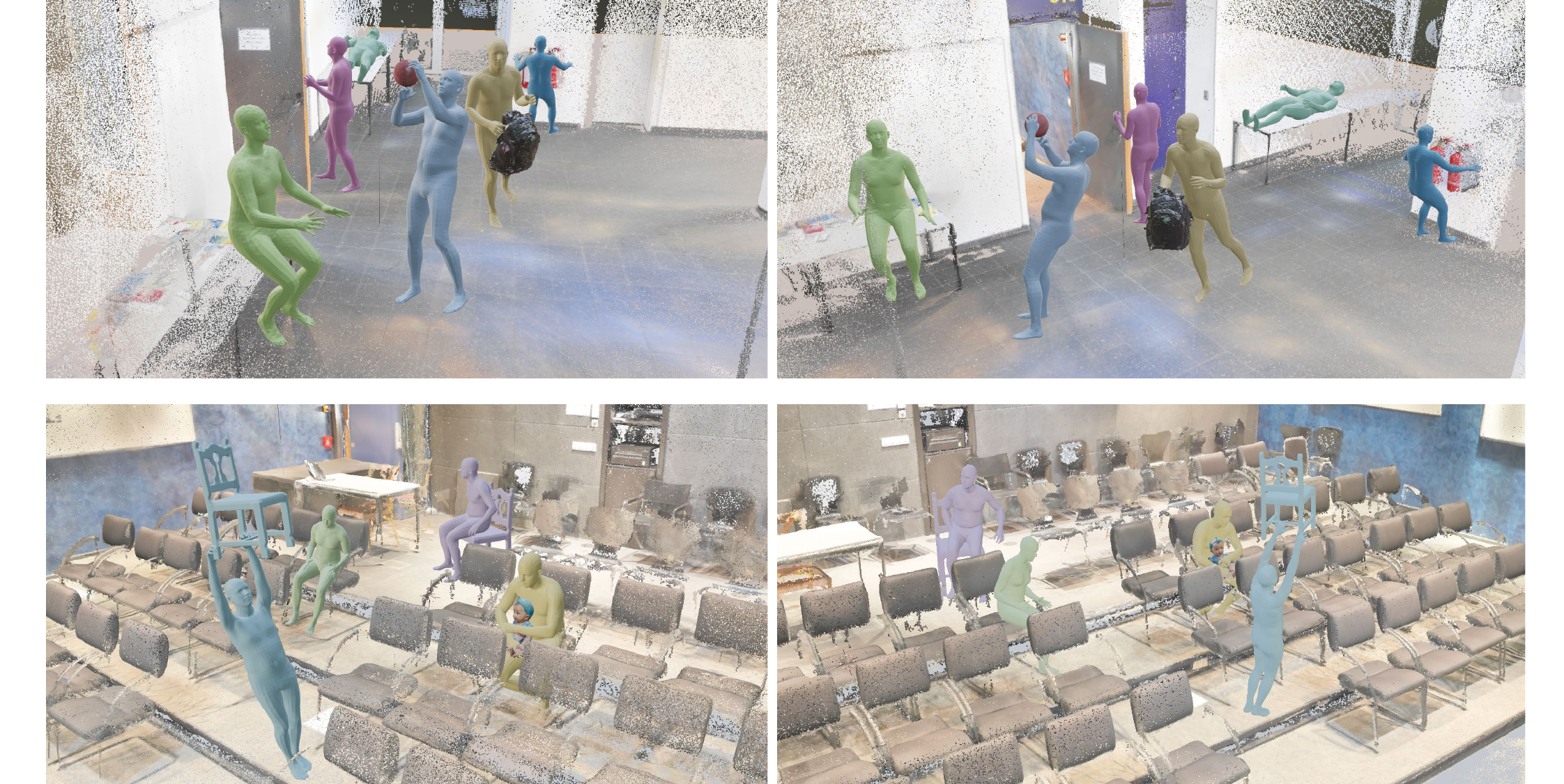}
    \caption{\textbf{Qualitative results of scene populating and scene interaction applications from different views.}} 
    \label{fig:scene}
    \vspace{-3mm}
\end{figure}

\noindent\textbf{Qualitative performance on Scene Populating.}
We demonstrate one possible way to populate scenes with \method. We first utilize the HPS dataset~\cite{HPS_dataset} to obtain the 3D scene and then integrate the generated HOI results into the scene by placing them in appropriate locations. By employing a ground loss and large language model, we ensure that the HOI results are naturally embedded within the scene without artifact issues such as object or human interpenetration. Furthermore, our method goes beyond simply populating HOI results into the scene, we also generate realistic interaction effects with existing objects in the environment. For example, \method can effectively generate interactions such as ``sitting on the table'' or ``opening a door'', as demonstrated in the \Cref{fig:scene}.

\subsection{Ablation Study}
\textbf{Ablation on Fine-grained Optimization}
In this study, we analyze the effect of fine-grained optimization terms as outlined in \Cref{sec:hoiopt} while keeping all other parameters constant.
The results shown in \Cref{fig:ablation-1}, clearly demonstrate that the fine-grained optimization significantly enhances the realism and contact plausibility of human-object interactions (right green panel) in the ``grasp'' prompt.
Without fine-grained optimization (left red panel), the alignment between the hand and table is noticeably inaccurate, resulting in unnatural poses where the hand appears to "float" near the table instead of properly grasping or holding it. In contrast, with fine-grained optimization enabled (right panel), the human hand is aligned with the table surface, resulting in a more realistic interaction. 
This improvement highlights the importance of fine-grained optimization in achieving natural contact and precise positioning, especially in actions involving detailed hand-object interactions.

\noindent\textbf{Ablation Study on LLM-guided Initialization}
We further ablate the LLM-guided initialization and the corresponding HOI refinement module (fine-grained HOI synthesis) to investigate the contribution of our design in leveraging language models. As shown in \Cref{tab:ablation_llm}, we use GPT-4V selection as the evaluation metric. 
Compared to the full method (Full), the variant without refinement (w/o Refine) exhibits a noticeable decline in performance. This drop is primarily due to the absence of fine-grained contact optimization, which is critical for generating the most accurate human-object interactions. 
In contrast, when the LLM-guided initialization is ablated (w/o LLM-init), we randomly initialize both the human and object positions, followed by coarse HOI optimization. This approach results in a significant decrease in selection scores, ranking lower than DreamHOI~\cite{zhu2024dreamhoi}. This finding emphasizes the importance of an informed initialization strategy in establishing a solid foundation for HOI optimization. 
The stark differences in the ablation study underscore the effectiveness of our design. The integration of LLM-guided initialization and refinement is vital for achieving superior results.

\section{Discussion and Limitation}
In this work, we demonstrate that our approach can generate fine-grained, realistic HOIs for open-set 3D objects based on text descriptions.
While our method shows promising results in improving contact precision and adapting to unseen objects, it still relies on 2D priors as its primary guidance. Consequently, advancements in multi-view consistent image generation or fully 4D generation techniques would likely enhance our approach, leading to more coherent and robust 3D interactions.
Additionally, physical plausibility is also an important application in recent applications, \eg, in humanoid manipulation. We use force closure as a physical constraint while a high-quality simulator could further evaluate the generated HOI.
Furthermore, our method currently uses the SMPL-H model to represent human body geometry and articulation. While this model captures a wide range of human poses and movements, it limits the adaptability of our framework to diverse interaction agents, particularly those with non-human anatomies. These agents often have unique skeletal structures and movement patterns that differ significantly from human biomechanics, which our current system is not fully equipped to process.
In future work, we aim to address these limitations by investigating multi-view or 4D generation for enhanced consistency and exploring more flexible models to represent a broader array.

\section{Conclusion}
This paper tackles the challenge of generating novel 3D HOI involving unseen objects—a task where current methods often struggle. To address this issue, we propose \method, a novel zero-shot synthesis method driven by text instructions and not reliant on training with existing 3D datasets. It leverages guidance from the large language model to distill human-object relations and employs the pre-trained 2D diffusion model to parse contact areas with open-set objects, circumventing the limitations of existing HOI datasets.




{
    \small
    \bibliographystyle{ieeenat_fullname}
    \bibliography{reference}

\begin{thebibliography}{67}
\providecommand{\natexlab}[1]{#1}
\providecommand{\url}[1]{\texttt{#1}}
\expandafter\ifx\csname urlstyle\endcsname\relax
  \providecommand{\doi}[1]{doi: #1}\else
  \providecommand{\doi}{doi: \begingroup \urlstyle{rm}\Url}\fi

\bibitem[Achiam et~al.(2023)Achiam, Adler, Agarwal, Ahmad, Akkaya, Aleman, Almeida, Altenschmidt, Altman, Anadkat, et~al.]{achiam2023gpt4}
Josh Achiam, Steven Adler, Sandhini Agarwal, Lama Ahmad, Ilge Akkaya, Florencia~Leoni Aleman, Diogo Almeida, Janko Altenschmidt, Sam Altman, Shyamal Anadkat, et~al.
\newblock Gpt-4 technical report.
\newblock \emph{arXiv preprint arXiv:2303.08774}, 2023.

\bibitem[Bhatnagar et~al.(2022)Bhatnagar, Xie, Petrov, Sminchisescu, Theobalt, and Pons-Moll]{BEHAVEDatasetMethod2022Bhatnagar}
Bharat~Lal Bhatnagar, Xianghui Xie, Ilya~A Petrov, Cristian Sminchisescu, Christian Theobalt, and Gerard Pons-Moll.
\newblock Behave: Dataset and method for tracking human object interactions.
\newblock In \emph{Conference on Computer Vision and Pattern Recognition (CVPR)}, 2022.

\bibitem[Bogo et~al.(2016)Bogo, Kanazawa, Lassner, Gehler, Romero, and Black]{KeepItSMPL2016Bogo}
Federica Bogo, Angjoo Kanazawa, Christoph Lassner, Peter Gehler, Javier Romero, and Michael~J. Black.
\newblock Keep {{It SMPL}}: {{Automatic Estimation}} of {{3D Human Pose}} and {{Shape}} from a {{Single Image}}.
\newblock In \emph{Computer {{Vision}} -- {{ECCV}} 2016}, 2016.

\bibitem[Cao et~al.(2024)Cao, Cao, Han, Shan, and Wong]{cao2024dreamavatar}
Yukang Cao, Yan-Pei Cao, Kai Han, Ying Shan, and Kwan-Yee~K Wong.
\newblock Dreamavatar: Text-and-shape guided 3d human avatar generation via diffusion models.
\newblock In \emph{Proceedings of the IEEE/CVF Conference on Computer Vision and Pattern Recognition}, pages 958--968, 2024.

\bibitem[Cao et~al.(2017)Cao, Simon, Wei, and Sheikh]{openpose}
Zhe Cao, Tomas Simon, Shih-En Wei, and Yaser Sheikh.
\newblock Realtime multi-person 2d pose estimation using part affinity fields.
\newblock In \emph{Conference on Computer Vision and Pattern Recognition (CVPR)}, 2017.

\bibitem[Chen et~al.(2023)Chen, Chen, Jiao, and Jia]{fantasia3D}
Rui Chen, Yongwei Chen, Ningxin Jiao, and Kui Jia.
\newblock Fantasia3d: Disentangling geometry and appearance for high-quality text-to-3d content creation.
\newblock In \emph{International Conference on Computer Vision (ICCV)}, 2023.

\bibitem[Chen et~al.(2024)Chen, Wang, Wu, Pan, Jia, and Liu]{chen2024comboverse}
Yongwei Chen, Tengfei Wang, Tong Wu, Xingang Pan, Kui Jia, and Ziwei Liu.
\newblock Comboverse: Compositional 3d assets creation using spatially-aware diffusion guidance.
\newblock In \emph{European Conference on Computer Vision}, pages 128--146. Springer, 2024.

\bibitem[Dai et~al.(2018)Dai, Majumdar, and Tedrake]{force_closure}
Hongkai Dai, Anirudha Majumdar, and Russ Tedrake.
\newblock Synthesis and optimization of force closure grasps via sequential semidefinite programming.
\newblock \emph{Robotics Research: Volume 1}, pages 285--305, 2018.

\bibitem[Dai et~al.(2024)Dai, Li, Sun, Huang, Ma, Huang, Xu, and Hu]{dai2024interfusion}
Sisi Dai, Wenhao Li, Haowen Sun, Haibin Huang, Chongyang Ma, Hui Huang, Kai Xu, and Ruizhen Hu.
\newblock Interfusion: Text-driven generation of 3d human-object interaction.
\newblock In \emph{European Conference on Computer Vision}, pages 18--35. Springer, 2024.

\bibitem[DeepMind(2023)]{google2023gemini}
Google DeepMind.
\newblock Gemini: A family of highly capable multimodal models.
\newblock \emph{arXiv preprint arXiv:2312.11805}, 2023.

\bibitem[Fuji~Tsang et~al.(2022)Fuji~Tsang, Shugrina, Lafleche, Takikawa, Wang, Loop, Chen, Jatavallabhula, Smith, Rozantsev, Perel, Shen, Gao, Fidler, State, Gorski, Xiang, Li, Li, and Lebaredian]{KaolinLibrary}
Clement Fuji~Tsang, Maria Shugrina, Jean~Francois Lafleche, Towaki Takikawa, Jiehan Wang, Charles Loop, Wenzheng Chen, Krishna~Murthy Jatavallabhula, Edward Smith, Artem Rozantsev, Or Perel, Tianchang Shen, Jun Gao, Sanja Fidler, Gavriel State, Jason Gorski, Tommy Xiang, Jianing Li, Michael Li, and Rev Lebaredian.
\newblock Kaolin: A pytorch library for accelerating 3d deep learning research.
\newblock \url{https://github.com/NVIDIAGameWorks/kaolin}, 2022.

\bibitem[Gao et~al.(2024)Gao, Liu, Chen, Geiger, and Sch{\"o}lkopf]{gao2024graphdreamer}
Gege Gao, Weiyang Liu, Anpei Chen, Andreas Geiger, and Bernhard Sch{\"o}lkopf.
\newblock Graphdreamer: Compositional 3d scene synthesis from scene graphs.
\newblock In \emph{Conference on Computer Vision and Pattern Recognition (CVPR)}, 2024.

\bibitem[Gong et~al.(2023)Gong, Foo, Fan, Ke, Rahmani, and Liu]{gong2023diffpose}
Jia Gong, Lin~Geng Foo, Zhipeng Fan, Qiuhong Ke, Hossein Rahmani, and Jun Liu.
\newblock Diffpose: {{Toward}} more reliable 3d pose estimation.
\newblock In \emph{Proceedings of the {{IEEE}}/{{CVF}} Conference on Computer Vision and Pattern Recognition}, pages 13041--13051, 2023.

\bibitem[Guzov et~al.(2021)Guzov, Mir, Sattler, and Pons-Moll]{HPS_dataset}
Vladimir Guzov, Aymen Mir, Torsten Sattler, and Gerard Pons-Moll.
\newblock Human poseitioning system (hps): 3d human pose estimation and self-localization in large scenes from body-mounted sensors.
\newblock In \emph{{IEEE} Conference on Computer Vision and Pattern Recognition (CVPR)}. {IEEE}, 2021.

\bibitem[Han and Joo(2023)]{han2023chorus}
Sookwan Han and Hanbyul Joo.
\newblock Chorus: Learning canonicalized 3d human-object spatial relations from unbounded synthesized images.
\newblock In \emph{International Conference on Computer Vision (ICCV)}, 2023.

\bibitem[Hassan et~al.(2019)Hassan, Choutas, Tzionas, and Black]{hassan2019resolving}
Mohamed Hassan, Vasileios Choutas, Dimitrios Tzionas, and Michael~J. Black.
\newblock Resolving {3D} human pose ambiguities with {3D} scene constraints.
\newblock In \emph{International Conference on Computer Vision (ICCV)}, 2019.

\bibitem[Hong et~al.(2022)Hong, Zhang, Pan, Cai, Yang, and Liu]{hong2022avatarclip}
Fangzhou Hong, Mingyuan Zhang, Liang Pan, Zhongang Cai, Lei Yang, and Ziwei Liu.
\newblock Avatarclip: Zero-shot text-driven generation and animation of 3d avatars.
\newblock \emph{arXiv preprint arXiv:2205.08535}, 2022.

\bibitem[Hu et~al.(2024)Hu, Hu, and Liu]{hu2024gauhuman}
Shoukang Hu, Tao Hu, and Ziwei Liu.
\newblock Gauhuman: {{Articulated}} gaussian splatting from monocular human videos.
\newblock In \emph{Proceedings of the {{IEEE}}/{{CVF}} Conference on Computer Vision and Pattern Recognition}, pages 20418--20431, 2024.

\bibitem[Huang et~al.(2018)Huang, Su, and Guibas]{manifold}
Jingwei Huang, Hao Su, and Leonidas Guibas.
\newblock Robust watertight manifold surface generation method for shapenet models.
\newblock \emph{arXiv preprint arXiv:1802.01698}, 2018.

\bibitem[Jiang et~al.(2023)Jiang, Liu, Cao, Cui, Zhang, Chen, Wang, Zhu, and Huang]{chairsdataset}
Nan Jiang, Tengyu Liu, Zhexuan Cao, Jieming Cui, Zhiyuan Zhang, Yixin Chen, He Wang, Yixin Zhu, and Siyuan Huang.
\newblock Full-body articulated human-object interaction.
\newblock In \emph{International Conference on Computer Vision (ICCV)}, 2023.

\bibitem[Jiang et~al.(2024)Jiang, Zhang, Li, Ma, Wang, Chen, Liu, Zhu, and Huang]{scalingdataset2024sjiang}
Nan Jiang, Zhiyuan Zhang, Hongjie Li, Xiaoxuan Ma, Zan Wang, Yixin Chen, Tengyu Liu, Yixin Zhu, and Siyuan Huang.
\newblock Scaling up dynamic human-scene interaction modeling.
\newblock In \emph{Conference on Computer Vision and Pattern Recognition (CVPR)}, 2024.

\bibitem[Kanazawa et~al.(2018)Kanazawa, Black, Jacobs, and Malik]{kanazawa2018end}
Angjoo Kanazawa, Michael~J Black, David~W Jacobs, and Jitendra Malik.
\newblock End-to-end recovery of human shape and pose.
\newblock In \emph{Proceedings of the {{IEEE}} Conference on Computer Vision and Pattern Recognition}, pages 7122--7131, 2018.

\bibitem[Karras(2012)]{Karras2012intersect}
Tero Karras.
\newblock Maximizing parallelism in the construction of bvhs, octrees, and k-d trees.
\newblock In \emph{Proceedings of ACM SIGGRAPH/Eurographics Conference on High-Performance Graphics}, 2012.

\bibitem[Kerbl et~al.(2023)Kerbl, Kopanas, Leimk{\"u}hler, and Drettakis]{kerbl20233d}
Bernhard Kerbl, Georgios Kopanas, Thomas Leimk{\"u}hler, and George Drettakis.
\newblock 3d gaussian splatting for real-time radiance field rendering.
\newblock \emph{ACM Transactions on Graphics (TOG)}, 42\penalty0 (4):\penalty0 139--1, 2023.

\bibitem[Kim et~al.(2024)Kim, Han, Kwon, et~al.]{coma}
Hyeonwoo Kim, Sookwan Han, Patrick Kwon, et~al.
\newblock Beyond the contact: Discovering comprehensive affordance for 3d objects from pre-trained 2d diffusion models.
\newblock In \emph{European Conference on Computer Vision (ECCV)}, 2024.

\bibitem[Li et~al.(2023)Li, Wu, and Liu]{omomo2023}
Jiaman Li, Jiajun Wu, and C~Karen Liu.
\newblock Object motion guided human motion synthesis.
\newblock \emph{ACM Transactions on Graphics (TOG)}, 42\penalty0 (6):\penalty0 1--11, 2023.

\bibitem[Li et~al.(2024)Li, Clegg, Mottaghi, Wu, Puig, and Liu]{ControllableHumanObjectInteraction2023Li}
Jiaman Li, Alexander Clegg, Roozbeh Mottaghi, Jiajun Wu, Xavier Puig, and C~Karen Liu.
\newblock Controllable human-object interaction synthesis.
\newblock In \emph{European Conference on Computer Vision (ECCV)}, 2024.

\bibitem[Li and Dai(2024)]{li2024genzi}
Lei Li and Angela Dai.
\newblock Genzi: {{Zero-shot}} 3d human-scene interaction generation.
\newblock In \emph{Proceedings of the {{IEEE}}/{{CVF}} Conference on Computer Vision and Pattern Recognition}, pages 20465--20474, 2024.

\bibitem[Lin et~al.(2023{\natexlab{a}})Lin, Gao, Tang, Takikawa, Zeng, Huang, Kreis, Fidler, Liu, and Lin]{magic3d}
Chen-Hsuan Lin, Jun Gao, Luming Tang, Towaki Takikawa, Xiaohui Zeng, Xun Huang, Karsten Kreis, Sanja Fidler, Ming-Yu Liu, and Tsung-Yi Lin.
\newblock Magic3d: High-resolution text-to-3d content creation.
\newblock In \emph{Conference on Computer Vision and Pattern Recognition (CVPR)}, 2023{\natexlab{a}}.

\bibitem[Lin et~al.(2023{\natexlab{b}})Lin, Zeng, Wang, Zhang, and Li]{lin2023one}
Jing Lin, Ailing Zeng, Haoqian Wang, Lei Zhang, and Yu Li.
\newblock One-stage 3d whole-body mesh recovery with component aware transformer.
\newblock In \emph{Proceedings of the {{IEEE}}/{{CVF}} Conference on Computer Vision and Pattern Recognition}, pages 21159--21168, 2023{\natexlab{b}}.

\bibitem[Liu et~al.(2023{\natexlab{a}})Liu, Wu, Van~Hoorick, Tokmakov, Zakharov, and Vondrick]{liu2023zero}
Ruoshi Liu, Rundi Wu, Basile Van~Hoorick, Pavel Tokmakov, Sergey Zakharov, and Carl Vondrick.
\newblock Zero-1-to-3: Zero-shot one image to 3d object.
\newblock In \emph{International Conference on Computer Vision (ICCV)}, 2023{\natexlab{a}}.

\bibitem[Liu et~al.(2021)Liu, Liu, Jiao, Zhu, and Zhu]{dfc}
Tengyu Liu, Zeyu Liu, Ziyuan Jiao, Yixin Zhu, and Song-Chun Zhu.
\newblock Synthesizing diverse and physically stable grasps with arbitrary hand structures using differentiable force closure estimator.
\newblock \emph{IEEE Robotics and Automation Letters (RA-L)}, 7\penalty0 (1):\penalty0 470--477, 2021.

\bibitem[Liu et~al.(2024)Liu, Zhan, Tang, Shan, Zeng, Lin, Liu, and Liu]{liu2024humangaussian}
Xian Liu, Xiaohang Zhan, Jiaxiang Tang, Ying Shan, Gang Zeng, Dahua Lin, Xihui Liu, and Ziwei Liu.
\newblock Humangaussian: Text-driven 3d human generation with gaussian splatting.
\newblock In \emph{Conference on Computer Vision and Pattern Recognition (CVPR)}, 2024.

\bibitem[Liu et~al.(2023{\natexlab{b}})Liu, Lin, Zeng, Long, Liu, Komura, and Wang]{liu2023syncdreamer}
Yuan Liu, Cheng Lin, Zijiao Zeng, Xiaoxiao Long, Lingjie Liu, Taku Komura, and Wenping Wang.
\newblock Syncdreamer: Generating multiview-consistent images from a single-view image.
\newblock In \emph{International Conference on Learning Representations (ICLR)}, 2023{\natexlab{b}}.

\bibitem[Long et~al.(2024)Long, Guo, Lin, Liu, Dou, Liu, Ma, Zhang, Habermann, Theobalt, et~al.]{long2024wonder3d}
Xiaoxiao Long, Yuan-Chen Guo, Cheng Lin, Yuan Liu, Zhiyang Dou, Lingjie Liu, Yuexin Ma, Song-Hai Zhang, Marc Habermann, Christian Theobalt, et~al.
\newblock Wonder3d: Single image to 3d using cross-domain diffusion.
\newblock In \emph{Conference on Computer Vision and Pattern Recognition (CVPR)}, 2024.

\bibitem[Loper et~al.(2015)Loper, Mahmood, Romero, Pons-Moll, and Black]{loper2015smpl}
Matthew Loper, Naureen Mahmood, Javier Romero, Gerard Pons-Moll, and Michael~J. Black.
\newblock {SMPL}: A skinned multi-person linear model.
\newblock \emph{ACM Transactions on Graphics (TOG)}, 34\penalty0 (6):\penalty0 248:1--248:16, 2015.

\bibitem[Mildenhall et~al.(2020)Mildenhall, Srinivasan, Tancik, Barron, Ramamoorthi, and Ng]{mildenhall2020nerf}
Ben Mildenhall, Pratul~P. Srinivasan, Matthew Tancik, Jonathan~T. Barron, Ravi Ramamoorthi, and Ren Ng.
\newblock {NeRF}: Representing scenes as neural radiance fields for view synthesis.
\newblock In \emph{European Conference on Computer Vision (ECCV)}, 2020.

\bibitem[Mo et~al.(2019)Mo, Zhu, Chang, Yi, Tripathi, Guibas, and Su]{mo2019partnet}
Kaichun Mo, Shilin Zhu, Angel~X Chang, Li Yi, Subarna Tripathi, Leonidas~J Guibas, and Hao Su.
\newblock Partnet: A large-scale benchmark for fine-grained and hierarchical part-level 3d object understanding.
\newblock In \emph{Proceedings of the IEEE/CVF conference on computer vision and pattern recognition}, pages 909--918, 2019.

\bibitem[OpenAI(2023)]{openai2023gpt35}
OpenAI.
\newblock Gpt-3.5.
\newblock \url{https://platform.openai.com/docs/models/gpt-3-5}, 2023.
\newblock Accessed: 2023-11-14.

\bibitem[Pavlakos et~al.(2019)Pavlakos, Choutas, Ghorbani, Bolkart, Osman, Tzionas, and Black]{SMPL-X}
Georgios Pavlakos, Vasileios Choutas, Nima Ghorbani, Timo Bolkart, Ahmed A.~A. Osman, Dimitrios Tzionas, and Michael~J. Black.
\newblock Expressive body capture: 3d hands, face, and body from a single image.
\newblock In \emph{Proceedings IEEE Conf. on Computer Vision and Pattern Recognition (CVPR)}, 2019.

\bibitem[Peng et~al.(2023)Peng, Xie, Wu, Jampani, Sun, and Jiang]{HOIDiffTextDrivenSynthesis2023Peng}
Xiaogang Peng, Yiming Xie, Zizhao Wu, Varun Jampani, Deqing Sun, and Huaizu Jiang.
\newblock Hoi-diff: Text-driven synthesis of 3d human-object interactions using diffusion models.
\newblock \emph{arXiv preprint arXiv:2312.06553}, 2023.

\bibitem[Podell et~al.(2023)Podell, English, Lacey, Blattmann, Dockhorn, M{\"u}ller, Penna, and Rombach]{podell2023sdxlimprovinglatentdiffusion}
Dustin Podell, Zion English, Kyle Lacey, Andreas Blattmann, Tim Dockhorn, Jonas M{\"u}ller, Joe Penna, and Robin Rombach.
\newblock Sdxl: Improving latent diffusion models for high-resolution image synthesis.
\newblock \emph{arXiv preprint arXiv:2307.01952}, 2023.

\bibitem[Poole et~al.(2023)Poole, Jain, Barron, and Mildenhall]{poole2022dreamfusion}
Ben Poole, Ajay Jain, Jonathan~T Barron, and Ben Mildenhall.
\newblock Dreamfusion: Text-to-3d using 2d diffusion.
\newblock In \emph{International Conference on Learning Representations (ICLR)}, 2023.

\bibitem[Radford et~al.(2021)Radford, Kim, Hallacy, Ramesh, Goh, Agarwal, Sastry, Amanda, Mishkin, Clark, Krueger, and Sutskever]{clip}
Alec Radford, JongWook Kim, Chris Hallacy, A. Ramesh, Gabriel Goh, Sandhini Agarwal, Girish Sastry, Askell Amanda, Pamela Mishkin, Jack Clark, Gretchen Krueger, and Ilya Sutskever.
\newblock Learning transferable visual models from natural language supervision.
\newblock \emph{Cornell University - arXiv,Cornell University - arXiv}, 2021.

\bibitem[Ramesh et~al.(2022)Ramesh, Dhariwal, Nichol, Chu, and Chen]{Ramesh_Dhariwal_Nichol_Chu_Chen}
Aditya Ramesh, Prafulla Dhariwal, Alex Nichol, Casey Chu, and Mark Chen.
\newblock Hierarchical text-conditional image generation with clip latents.
\newblock \emph{arXiv preprint arXiv:2204.06125}, 2022.

\bibitem[Rombach et~al.(2022{\natexlab{a}})Rombach, Blattmann, Lorenz, Esser, and Ommer]{rombach2022highresolutiondiffusion}
Robin Rombach, Andreas Blattmann, Dominik Lorenz, Patrick Esser, and Bj{\"o}rn Ommer.
\newblock High-resolution image synthesis with latent diffusion models.
\newblock In \emph{Conference on Computer Vision and Pattern Recognition (CVPR)}, 2022{\natexlab{a}}.

\bibitem[Rombach et~al.(2022{\natexlab{b}})Rombach, Blattmann, Lorenz, Esser, and Ommer]{stablediffusion_2022_Rombach}
Robin Rombach, Andreas Blattmann, Dominik Lorenz, Patrick Esser, and Bj\"orn Ommer.
\newblock High-resolution image synthesis with latent diffusion models.
\newblock In \emph{Proceedings of the IEEE/CVF Conference on Computer Vision and Pattern Recognition (CVPR)}, pages 10684--10695, 2022{\natexlab{b}}.

\bibitem[Romero et~al.(2017)Romero, Tzionas, and Black]{MANO2017}
Javier Romero, Dimitrios Tzionas, and Michael~J Black.
\newblock Embodied hands: modeling and capturing hands and bodies together.
\newblock \emph{ACM Transactions on Graphics (TOG)}, 36\penalty0 (6):\penalty0 1--17, 2017.

\bibitem[Ruiz et~al.(2023)Ruiz, Li, Jampani, Pritch, Rubinstein, and Aberman]{ruiz2023dreambooth}
Nataniel Ruiz, Yuanzhen Li, Varun Jampani, Yael Pritch, Michael Rubinstein, and Kfir Aberman.
\newblock Dreambooth: Fine tuning text-to-image diffusion models for subject-driven generation.
\newblock In \emph{Conference on Computer Vision and Pattern Recognition (CVPR)}, 2023.

\bibitem[Saharia et~al.(2022)Saharia, Chan, Saxena, Li, Whang, Denton, Ghasemipour, Gontijo~Lopes, Karagol~Ayan, Salimans, et~al.]{Photorealisticdiffusion}
Chitwan Saharia, William Chan, Saurabh Saxena, Lala Li, Jay Whang, Emily~L Denton, Kamyar Ghasemipour, Raphael Gontijo~Lopes, Burcu Karagol~Ayan, Tim Salimans, et~al.
\newblock Photorealistic text-to-image diffusion models with deep language understanding.
\newblock In \emph{Advances in Neural Information Processing Systems (NeurIPS)}, 2022.

\bibitem[Shi et~al.(2023{\natexlab{a}})Shi, Chen, Zhang, Liu, Xu, Wei, Chen, Zeng, and Su]{shi2023zero123++}
Ruoxi Shi, Hansheng Chen, Zhuoyang Zhang, Minghua Liu, Chao Xu, Xinyue Wei, Linghao Chen, Chong Zeng, and Hao Su.
\newblock Zero123++: a single image to consistent multi-view diffusion base model.
\newblock \emph{arXiv preprint arXiv:2310.15110}, 2023{\natexlab{a}}.

\bibitem[Shi et~al.(2023{\natexlab{b}})Shi, Wang, Ye, Long, Li, and Yang]{shi2023mvdream}
Yichun Shi, Peng Wang, Jianglong Ye, Mai Long, Kejie Li, and Xiao Yang.
\newblock Mvdream: Multi-view diffusion for 3d generation.
\newblock \emph{arXiv preprint arXiv:2308.16512}, 2023{\natexlab{b}}.

\bibitem[Taheri et~al.(2020)Taheri, Ghorbani, Black, and Tzionas]{grabdataset2020}
Omid Taheri, Nima Ghorbani, Michael~J Black, and Dimitrios Tzionas.
\newblock Grab: A dataset of whole-body human grasping of objects.
\newblock In \emph{European Conference on Computer Vision (ECCV)}, 2020.

\bibitem[Tripathi et~al.(2023)Tripathi, Chatterjee, Passy, Yi, Tzionas, and Black]{DECODenseEstimation2023Tripathi}
Shashank Tripathi, Agniv Chatterjee, Jean-Claude Passy, Hongwei Yi, Dimitrios Tzionas, and Michael~J Black.
\newblock Deco: Dense estimation of 3d human-scene contact in the wild.
\newblock In \emph{International Conference on Computer Vision (ICCV)}, 2023.

\bibitem[Wang et~al.(2022)Wang, Li, Kuo, Kocabas, Aksan, and Hilliges]{wang2022reconstructing}
Xi Wang, Gen Li, Yen-Ling Kuo, Muhammed Kocabas, Emre Aksan, and Otmar Hilliges.
\newblock Reconstructing action-conditioned human-object interactions using commonsense knowledge priors.
\newblock In \emph{International Conference on 3D Vision (3DV)}, 2022.

\bibitem[Wang et~al.(2024)Wang, Lu, Wang, Bao, Li, Su, and Zhu]{prolificDreamer}
Zhengyi Wang, Cheng Lu, Yikai Wang, Fan Bao, Chongxuan Li, Hang Su, and Jun Zhu.
\newblock Prolificdreamer: High-fidelity and diverse text-to-3d generation with variational score distillation.
\newblock In \emph{Advances in Neural Information Processing Systems (NeurIPS)}, 2024.

\bibitem[Weng et~al.(2022)Weng, Curless, Srinivasan, Barron, and Kemelmacher-Shlizerman]{HumanNeRF2022Weng}
Chung-Yi Weng, Brian Curless, Pratul~P. Srinivasan, Jonathan~T. Barron, and Ira Kemelmacher-Shlizerman.
\newblock Humannerf: Free-viewpoint rendering of moving people from monocular video.
\newblock In \emph{Proceedings of the IEEE/CVF Conference on Computer Vision and Pattern Recognition}, pages 16210--16220, 2022.

\bibitem[Xu et~al.(2023)Xu, Li, Wang, and Gui]{xu2023interdiff}
Sirui Xu, Zhengyuan Li, Yu-Xiong Wang, and Liang-Yan Gui.
\newblock {InterDiff}: Generating 3d human-object interactions with physics-informed diffusion.
\newblock In \emph{International Conference on Computer Vision (ICCV)}, 2023.

\bibitem[Xu et~al.(2024)Xu, Wang, Gui, et~al.]{xu2024interdreamer}
Sirui Xu, Yu-Xiong Wang, Liangyan Gui, et~al.
\newblock Interdreamer: Zero-shot text to 3d dynamic human-object interaction.
\newblock \emph{Advances in Neural Information Processing Systems}, 37:\penalty0 52858--52890, 2024.

\bibitem[Yi et~al.(2025)Yi, Thies, Black, Peng, and Rempe]{GeneratingHumanInteraction2024Yi}
Hongwei Yi, Justus Thies, Michael~J Black, Xue~Bin Peng, and Davis Rempe.
\newblock Generating human interaction motions in scenes with text control.
\newblock In \emph{European Conference on Computer Vision (ECCV)}, 2025.

\bibitem[Yin et~al.(2024)Yin, Cai, Wang, Wang, Wei, Mei, Xiao, Yang, Sun, Yamashita, et~al.]{yin2024whac}
Wanqi Yin, Zhongang Cai, Ruisi Wang, Fanzhou Wang, Chen Wei, Haiyi Mei, Weiye Xiao, Zhitao Yang, Qingping Sun, Atsushi Yamashita, et~al.
\newblock Whac: {{World-grounded}} humans and cameras.
\newblock In \emph{European Conference on Computer Vision}, pages 20--37, 2024.

\bibitem[Zhang et~al.(2023)Zhang, Tian, Zhang, Li, An, Sun, and Liu]{PyMAFX2023Zhang}
Hongwen Zhang, Yating Tian, Yuxiang Zhang, Mengcheng Li, Liang An, Zhenan Sun, and Yebin Liu.
\newblock Pymaf-x: Towards well-aligned full-body model regression from monocular images.
\newblock \emph{IEEE Transactions on Pattern Analysis and Machine Intelligence}, 45:\penalty0 12287--12303, 2023.

\bibitem[Zhang et~al.(2022)Zhang, Tu, Yang, Chen, and Yuan]{MixSTE2022Zhang}
Jinlu Zhang, Zhigang Tu, Jianyu Yang, Yujin Chen, and Junsong Yuan.
\newblock Mixste: Seq2seq mixed spatio-temporal encoder for 3d human pose estimation in video.
\newblock In \emph{Proceedings of the IEEE/CVF Conference on Computer Vision and Pattern Recognition (CVPR)}, pages 13232--13242, 2022.

\bibitem[Zhang et~al.(2020)Zhang, Pepose, Joo, Ramanan, Malik, and Kanazawa]{zhang2020phosa}
Jason~Y. Zhang, Sam Pepose, Hanbyul Joo, Deva Ramanan, Jitendra Malik, and Angjoo Kanazawa.
\newblock Perceiving 3d human-object spatial arrangements from a single image in the wild.
\newblock In \emph{European Conference on Computer Vision (ECCV)}, 2020.

\bibitem[Zhao et~al.(2022)Zhao, Yang, Zhang, Lin, Zhang, Yu, and Xu]{Humannerf2022Zhao}
Fuqiang Zhao, Wei Yang, Jiakai Zhang, Pei Lin, Yingliang Zhang, Jingyi Yu, and Lan Xu.
\newblock Humannerf: Efficiently generated human radiance field from sparse inputs.
\newblock In \emph{Proceedings of the IEEE/CVF Conference on Computer Vision and Pattern Recognition}, pages 7743--7753, 2022.

\bibitem[Zheng et~al.(2021)Zheng, Yu, Liu, and Dai]{zheng2021pamir}
Zerong Zheng, Tao Yu, Yebin Liu, and Qionghai Dai.
\newblock Pamir: Parametric model-conditioned implicit representation for image-based human reconstruction.
\newblock \emph{IEEE transactions on pattern analysis and machine intelligence}, 44\penalty0 (6):\penalty0 3170--3184, 2021.

\bibitem[Zhu et~al.(2024)Zhu, Li, and Jakab]{zhu2024dreamhoi}
Thomas~Hanwen Zhu, Ruining Li, and Tomas Jakab.
\newblock Dreamhoi: Subject-driven generation of 3d human-object interactions with diffusion priors.
\newblock \emph{arXiv preprint arXiv:2409.08278}, 2024.

\end{thebibliography}
}


\end{document}